\documentclass[onecolumn, a4paper,journal,10pt]{IEEEtran}
\usepackage{amsfonts}
\usepackage{graphicx, epsfig,amsmath,amssymb,latexsym,graphics,float}
\usepackage{setspace,subfig}
\usepackage{citesort}
\usepackage{float}
\usepackage[ruled,vlined,linesnumbered]{algorithm2e}
\include{IEEEtran.bib}

\begin{document}

\title{Consistent Relative Confidence and Label-Free Model Selection for Convolutional Neural Networks}
\author{Bin~Liu\\
Research Center for Applied Mathematics and Machine Intelligence\\
Zhejiang Lab, Hangzhou, 311100 China\\
E-mail: bins@ieee.org
}
\maketitle

\begin{abstract}
In this paper, we are concerned with image classification with deep convolutional neural networks (CNNs). We focus on the following question: given a set of candidate CNN models, how to select the right one with the best generalization property for the current task? Current model selection methods all require access to a batch of labeled data for computing a pre-specified performance metric, such as the cross-entropy loss, the classification error rate and the negative log-likelihood. In many practical cases, labels are not available in time as labeling itself is a time-consuming and expensive task. To this end, we propose an approach to CNN model selection using only unlabeled data. We develop this method based on a principle termed consistent relative confidence. Experimental results on benchmark datasets demonstrate the effectiveness and efficiency of our method.
\end{abstract}

\IEEEkeywords% begin{keywords}
convolutional neural network, consistent relative confidence, label-free model selection, image classification
%\end{keywords}

\section{Introduction}
In the last decade, CNNs have become the dominant modeling and algorithmic framework for various computer vision
tasks such as image classification \cite{lecun2015deep,krizhevsky2012imagenet}. In this paper, we are concerned with a practical issue related to CNNs' applications: given a set of candidate CNN models, how to select the right one that has the best generalization performance for the current task? For ease of presentation, we only focus on the image classification task here, while our method can be generalized to other CNN application scenarios.

The model selection problem appears in different cases. For example, one may have several candidate models pre-trained from different source domains in a transfer learning case. Then, which is the right one to choose for transferring knowledge to the target domain? In the context of continual or life-long learning \cite{parisi2019continual}, several candidate models may be available from previous tasks. Again, one needs to select the right one for the forthcoming task. Edge learning in the context of cloud computing, see e.g., \cite{chen2019deep}, also calls for lightweight learning methods such as selecting one model for use at the edge node from a set of candidate models pre-trained at the server side.

Broadly speaking, any supervised training or fine-tuning operation on a model can be regarded as a model selection process, where the candidate models span over the whole parameter space. The optimized parameters resulting from the training or fine-tuning process define the model finally selected. In this generalized model selection problem, the number of candidate models is infinite, while, here we only consider conventional model selection problems, wherein the number of candidate models is limited.

Current approaches to neural network model selection, e.g., hypothesis testing, information criteria, cross validation, all require access to the labels of testing data, which are used for computing a pre-specified performance metric, such as the cross-entropy (CE) loss, the classification error rate (ER), and the negative log-likelihood (NLL) \cite{anders1999model}.

But, \emph{can we do accurate model selection based solely on unlabeled data?}

This question motivates this work. We believe that answering this question is crucially important for many practical scenarios. For example, some scenarios require predictions to be made timely, while there is no time to train a new model from scratch, yet some candidate models built based on historical data are available for use. It is natural to ask: which is the right one to select? Further, labeled data for model selection are not available, as labeling itself is a time-consuming and expensive task. We refer to the above problem as label-free model selection (LFMS). LFMS is in spirit related to self-supervised learning, see e.g., \cite{jing2020self}, and unsupervised clustering, see e.g., \cite{xie2016unsupervised,caron2018deep}, while they are different in the problem setting. Compared with self-supervised learning or unsupervised clustering, LFMS is characterized by an underlying assumption present in its problem setting, namely, a set of pre-trained models is available.
%and the emphasis is on how to evaluate and compare such models, while self-supervised learning or unsupervised clustering focuses on the modeling issue itself.

In this paper, we propose an extremely simple-to-implement yet highly effective and efficient approach to LFMS (Section \ref{sec:model_sel}). We develop this approach based on a principle termed consistent relative confidence (Section \ref{sec:crc}). To our knowledge, this work is the first to introduce this principle in the literature. We test the performance of our approach via experiments in Section \ref{sec:experiments} and conclude the paper in Section \ref{sec:discussion}.
%been the dominated Despite that deep learning methods model selection \cite{ghosh2019model}. Unsupervised learning of CNN \cite{liu2018unsupervised,wang2015unsupervised,dosovitskiy2014discriminative}.
\section{Consistent Relative Confidence}\label{sec:crc}
Here we present the consistent relative confidence (CRC) principle.

To begin with, let consider a thought experiment, in which some volunteers are asked to finish an image classification task. They all behave under the assumption of perfect rationality \cite{steele2004understanding}, which indicate that they shall make decisions as accurate as possible according to their experience previously collected. Consider the following question: is there a positive relationship between a volunteer's confidence and the quality of his (her) decision? The term confidence reflects an internal judgment of the volunteer on the probability of his (her) decision being correct. The work in \cite{power1994impacts} reports a positive relationship between confidence and correctness, while the work in \cite{landsbergen1997decision} negates it.

Confidence and correctness are not the same, while, we find that there is a strong positive relationship between consistent relative confidence (CRC) and correctness. Specifically, we take a ``rational" pre-trained CNN model as a volunteer for an image classification task. We find that, if there is one volunteer that behaves more confidently than the others in a consistent way over multiple testing images to be classified, then its decision is likely to be accurate. Such a phenomenon suggests that we can select the model that has the largest CRC score for use (the definition of the CRC score is deferred to Section \ref{sec:model_sel}). Such a model (volunteer) selection procedure only draws information from each model's internal judgment, while requires no label information. Therefore, it is a label-free approach to model selection. Experimental results show that our approach could make decisions as accurate as methods that depend on data labels.

To summarize, the CRC principle, which we find and propose here, is: among a group of \textbf{rational} volunteers, if there is one that acts \textbf{more confidently} than the others in a \textbf{consistent} way, then his (or her) decision is more accurate than the others'. In another word, while confidence and correctness are not the same, here we build a ``bridge" that connects confidence to correctness, whose building elements are relativity, consistency, and rationality.

The basic idea underlying our model selection approach is to find out the expert (resp. the right model) among the volunteers (resp. the candidate models) based on the CRC principle, namely to select the one with the largest CRC score.
%Certainly he (or she) will be consistently much more confident in classifying images during this task. Such CRC principle states one phenomenon that  there is a volunteer who is much more expert than the others in image classification.
\section{CRC based Label-Free CNN Model Selection}\label{sec:model_sel}
Here we describe the proposed CRC based label-free approach to CNN model selection.

Let focus on an image classification task. Suppose that we have at hand $K$ ``rational" pre-trained models $M_1, M_2, \ldots, M_K$, and $N$ images $I_1, I_2, \ldots, I_N$ to be categorized. Note that all candidate models being ``rational" is the precondition or constraint that makes our approach work. Such a constraint ensures that every candidate model is built delicately, not arbitrarily.

Denote the number of classes as $C$. The last layer of the CNN architecture is fixed to be softmax. Given an image $I_n$ as the input of $M_k$, $n=1,\ldots,N, k=1,\ldots,K$, one forward propagation running of $M_k$ outputs a probability vector $[p_{1,n,k}, p_{2,n,k}, \ldots, p_{C,n,k}]$ that satisfies $0\leq p_{c,n,k}\leq 1, c=1,\ldots,C$ and $\sum_{c=1}^Cp_{c,n,k}=1$.
Define the ``confidence" of $M_k$ on image $I_n$ to be
\begin{equation}\label{def:rc}
\mathcal{C}_{k,n}=\max\{p_{1,n,k}, p_{2,n,k}, \ldots, p_{C,n,k}\}.
\end{equation}
Let
\begin{eqnarray}
\mathcal{MC}_{k}&\triangleq & \frac{1}{N}\sum_{n=1}^{N}\mathcal{C}_{k,n}, \\
\mathcal{SC}_{k}&\triangleq & \sqrt{\frac{1}{N-1}\sum_{n=1}^{N}(\mathcal{C}_{k,n}-\mathcal{MC}_{k})^2}.
\end{eqnarray}
Then we define respectively the lower confidence bound (LCB) and the CRC score of $M_k$ on images $\{I_1, I_2, \ldots, I_N\}$ as
\begin{equation}\label{def:lcb}
\mathcal{LCB}_{k}=\mathcal{MC}_{k}-\mathcal{SC}_{k},
\end{equation}
and
\begin{equation}\label{def:crc_score}
\mathcal{S}_k=\mathcal{LCB}_k-\max_{i\in\{1:K\}\backslash k}\mathcal{LCB}_{i},
\end{equation}
where $\{1:K\}\backslash k\triangleq[1,\ldots,k-1,k+1,\ldots,K]$.

Based on the above setting, the proposed model selection procedure is to identify the index $k^{\star}$ that satisfies
\begin{equation}\label{eqn:ms}
k^{\star}=\arg\max_{k=1:K}\mathcal{S}_k=\arg\max_{k=1:K}\mathcal{LCB}_{k},
\end{equation}
and then select $M_{k^{\star}}$ for use.
The maximization operation in Eqn.(\ref{eqn:ms}) guarantees that the model finally selected, namely $M_{k^{\star}}$, is relatively more confident than the other models in a consistent way. The consistency property is resulted from the application of the LCB in Eqn.(\ref{eqn:ms}). Specifically, the smaller is $\mathcal{SC}_{k^{\star}}$, the larger is the extent to which $M_{k^{\star}}$'s confidence is consistently greater than the others'.

The CRC score of the selected model, namely $\mathcal{S}_{k^{\star}}$, takes a value between 0 and 1, and scores of the other models are negative. The CRC score of $\mathcal{S}_{k^{\star}}$ acts as an internal judgment of the reliability (or accuracy) of the model selection result. The closer it is to 1, the more reliable (or accurate) is the prediction given by $M_{k^{\star}}$, and vice versa. Note that no data label is involved in the above operations.
\section{Experiments}\label{sec:experiments}
Here we present the experimental results. The purpose of the experiment is twofold: to check whether the CRC principle and the model selection approach described in Section \ref{sec:model_sel} are effective and to evaluate our method's efficiency.
\subsection{Experimental setting}\label{sec:exp_setting}
We test our method on two widely used datasets, MNIST and FasionMNIST. Each dataset contains 60,000 training samples, 10,000 testing samples, and 10 image classes. Each data sample consists of an image and an associated label.

For each original dataset, we generate 4 related datasets associated with it. Each generated dataset is associated with a specific image processing operation, which is performed on every image included in the original dataset. That means each generated dataset is of the same size as the original one. Every image in a generated dataset is assigned with the same label as its counterpart in the original dataset. The image operations we use are:
\begin{itemize}
  \item Operation 1: image filtering with a rotationally symmetric Gaussian lowpass filter of size 7 with standard deviation 1;
  \item Operation 2: image filtering with a 3-by-3 filter whose shape approximates that of the 2D Laplacian operator;
  \item Operation 3: adding Gaussian white noise with mean 0 and variance 0.02 to each image;
  \item Operation 4: adding Gaussian white noise with mean 0 and variance 0.1 to each image.
\end{itemize}
\begin{figure}[!htb]
\centering
\includegraphics[width=1.8in,height=1.5in]{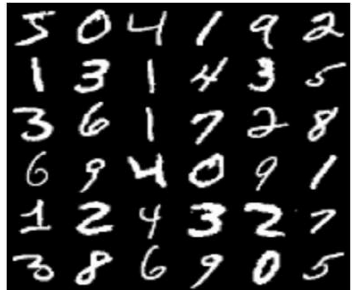}%[width=1.5in,height=1.25in]{Figs_tables/mnist_ori.eps}
\caption{Image samples drawn from the MNIST dataset.}
\label{fig:mnist}
\end{figure}
\begin{figure}[!htb]
\centering
\includegraphics[width=2.8in,height=2.8in]{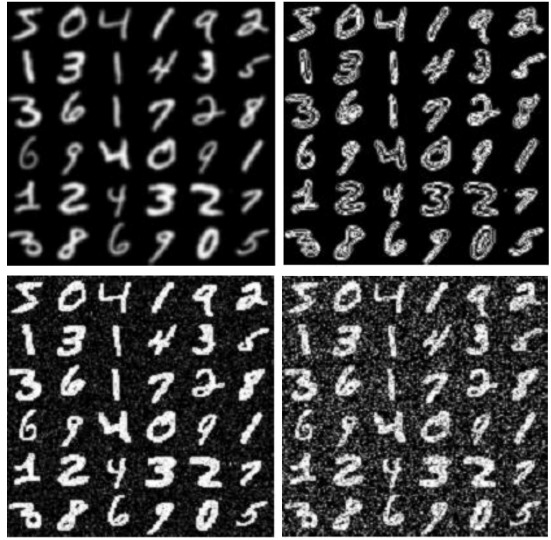}%[width=2.8in,height=2.8in]{Figs_tables/mnist_aug.eps}
\caption{Counterpart images of those shown in Fig.\ref{fig:mnist}. The 4 sub-figures correspond to 4 different image processing operations, respectively. See the text in subsection \ref{sec:exp_setting} for details on the operations.}
\label{fig:mnist_aug}
\end{figure}

\begin{figure}[!htb]
\centering
\includegraphics[width=1.8in,height=1.5in]{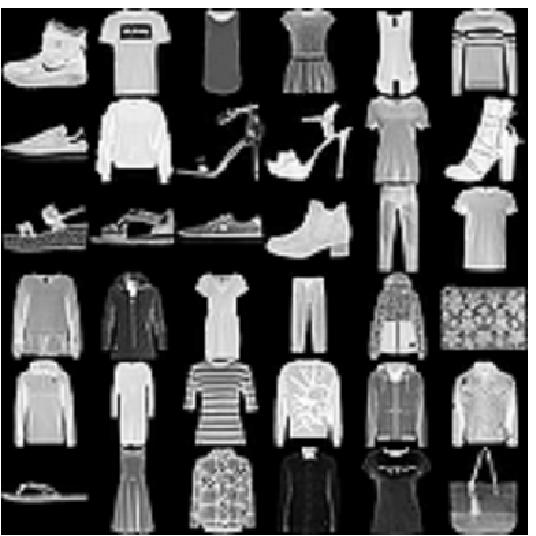}
\caption{Image samples drawn from the FasionMNIST dataset.}
\label{fig:fmnist}
\end{figure}

\begin{figure}[!htb]
\centering
\includegraphics[width=2.8in,height=2.8in]{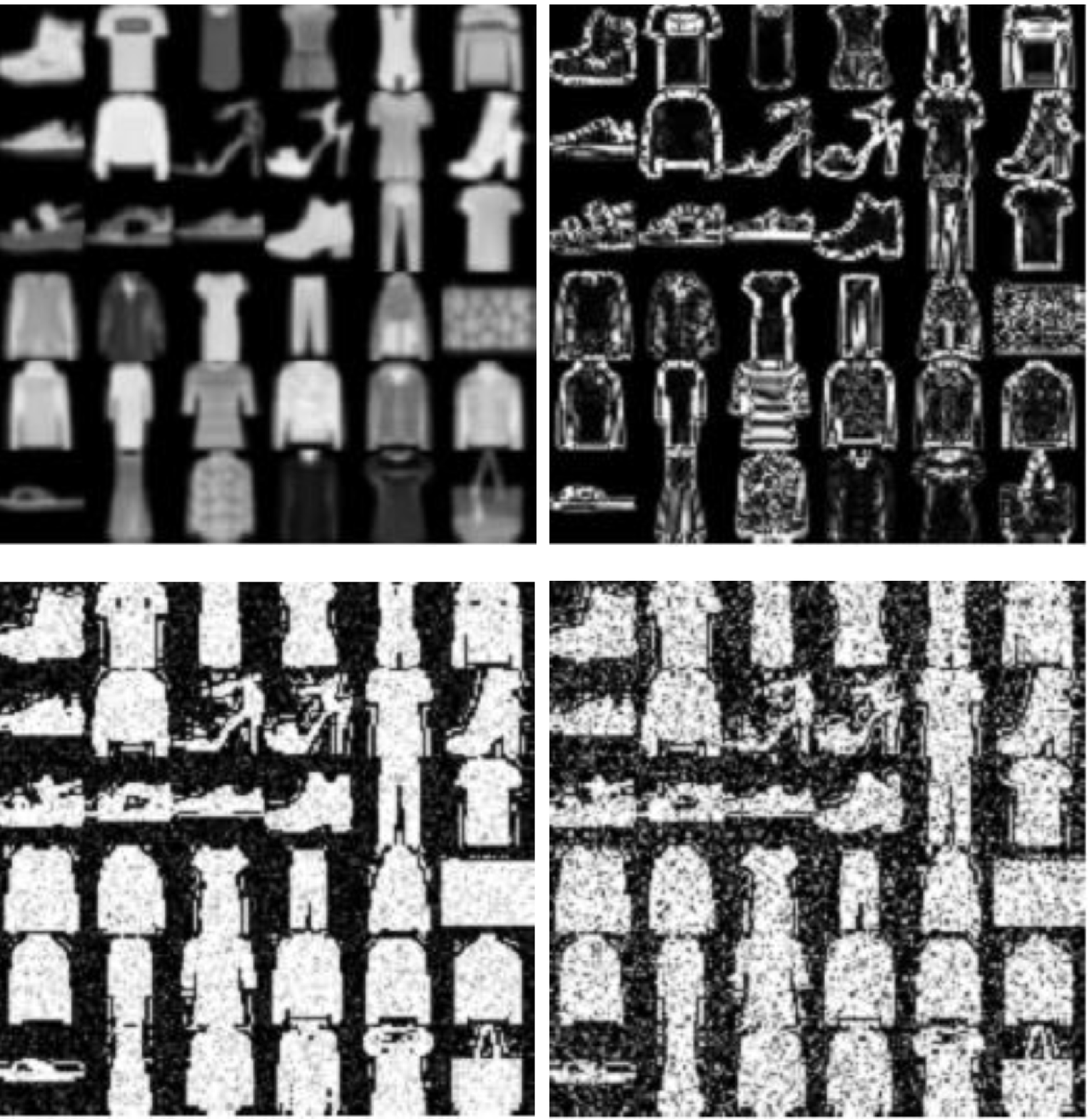}
\caption{Counterpart images of those shown in Fig.\ref{fig:fmnist}. The 4 sub-figures correspond to 4 different image processing operations, respectively. See the text in subsection \ref{sec:exp_setting} for details on the operations.}
\label{fig:fmnist_aug}
\end{figure}

See Figs.\ref{fig:mnist}-\ref{fig:fmnist_aug} for several original images and their counterparts in the generated datasets.
Denote the original dataset MNIST by MNIST-D0 and the newly generated datasets by MNIST-D$i$, $i=1,\ldots,4$. MNIST-D$i$ is generated with Operation $i, i=1,\ldots,4$. We obtain FasionMNIST-D$i$, $i=0,\ldots,4$, in the same way. Then we train one CNN model based on each dataset. Denote the resulting model associated with the $i$th dataset by $M_i$, $i=0,\ldots,4$. Only the training samples included in each dataset are used for model training. The remaining testing samples are used for performance evaluation of the involved model selection methods.
\begin{table*}[htbp] \small
\caption{Performance metrics for model selection based on $10^4$ testing images for each dataset. The models selected by each method are bolded. For the dataset MNIST-D$i$ (or FasionMNIST-D$i$), the associated ground truth model is $M_i$, $i=0,1,\ldots,4$.}
\label{table:stat_test}
\begin{center}
\begin{tabular}{c|c|c} %1
\hline
 & MNIST-D0 & MNIST-D1\\
\begin{tabular}{c} %2
 \\ \hline CRC \\ ER \\ CE
\end{tabular} %2
& \begin{tabular}{ccccc} %3
$M_0$ & $M_1$ & $M_2$ & $M_3$ & $M_4$ \\ \hline
\textbf{0.004} & -0.004 & -0.06 & -0.02 & -0.02 \\
\textbf{0.0155}  & 0.0161   & 0.0500 & 0.0160  & 0.0198 \\
\textbf{0.04} & 0.06 & 0.19 & 0.06  & 0.08
\end{tabular} %3
& \begin{tabular}{ccccc} %4
$M_0$ & $M_1$ & $M_2$ & $M_3$ & $M_4$ \\ \hline
-0.1001 & \textbf{0.068} & -0.068  & -0.0717   & -0.068 \\
0.0189 & \textbf{0.0122} & 0.0418 & 0.0204 & 0.0255 \\
0.087 & \textbf{0.0411} & 0.1305 & 0.0715  & 0.0914
\end{tabular} %4
\\ \hline
 & MNIST-D2 & MNIST-D3\\
\begin{tabular}{c} %2
  \\ \hline CRC \\ ER \\ CE
\end{tabular} %2
& \begin{tabular}{ccccc} %3
$M_0$ & $M_1$ & $M_2$ & $M_3$ & $M_4$ \\ \hline
-0.2534 & -0.1297 & \textbf{0.1297}  & -0.2172 & -0.2144 \\
0.1059 & 0.1490 & \textbf{0.0222} & 0.1052 & 0.1186 \\
0.6459 & 0.6509 & \textbf{0.0731} & 0.3504  & 0.3985
\end{tabular} %3
& \begin{tabular}{ccccc} %4
$M_0$ & $M_1$ & $M_2$ & $M_3$ & $M_4$ \\ \hline
-0.0488 & -0.004 & -0.0155  & \textbf{0.0011} & -0.0011 \\
0.0202 & 0.0192 & 0.0333 & \textbf{0.0161} & 0.0174 \\
0.082 & 0.0741 & 0.1369 & \textbf{0.0502}  & 0.0653
\end{tabular} %4
\\ \hline
 & MNIST-D4 & FasionMNIST-D0\\
\begin{tabular}{c} %2
  \\ \hline CRC \\ ER \\ CE
\end{tabular} %2
& \begin{tabular}{ccccc} %3
$M_0$ & $M_1$ & $M_2$ & $M_3$ & $M_4$ \\ \hline
-0.201 & -0.0107 & -0.0493  & -0.0234 & \textbf{0.0107} \\
0.0635 & 0.0312 & 0.0502 & \textbf{0.0205} & 0.0229 \\
0.2437 & 0.1044 & 0.1724 & \textbf{0.068}  & 0.0771
\end{tabular} %3
& \begin{tabular}{ccccc} %4
$M_0$ & $M_1$ & $M_2$ & $M_3$ & $M_4$ \\ \hline
-0.0214 & \textbf{0.0214} & -0.0861 & -0.1499 & -0.1178 \\
\textbf{0.0890} & 0.1550  & 0.4222   & 0.2294 & 0.2321 \\
\textbf{0.3185} & 0.4694 & 1.5092 & 0.628  & 0.7013
\end{tabular} %4
\\ \hline
 & FasionMNIST-D1 & FasionMNIST-D2\\
\begin{tabular}{c} %2
  \\ \hline CRC \\ ER \\ CE
\end{tabular} %2
& \begin{tabular}{ccccc} %3
$M_0$ & $M_1$ & $M_2$ & $M_3$ & $M_4$ \\ \hline
-0.073 & \textbf{0.073} & -0.0767 & -0.1685 & -0.1612 \\
0.1389 & \textbf{0.1346} & 0.3632   & 0.2489 & 0.2850 \\
0.4119 & \textbf{0.3763} & 0.9824 & 0.6535  & 0.7628
\end{tabular} %3
& \begin{tabular}{ccccc} %4
$M_0$ & $M_1$ & $M_2$ & $M_3$ & $M_4$ \\ \hline
-0.2005 & -0.2744 & \textbf{0.2005} & -0.3855 & -0.3016 \\
0.4536 & 0.5887   & \textbf{0.1208}   & 0.6826 & 0.6857 \\
2.0528 & 2.0292 & \textbf{0.3299} & 2.2149  & 2.8310
\end{tabular} %4
\\ \hline
 & FasionMNIST-D3 & FasionMNIST-D4\\
\begin{tabular}{c} %2
  \\ \hline CRC \\ ER \\ CE
\end{tabular} %2
& \begin{tabular}{ccccc} %3
$M_0$ & $M_1$ & $M_2$ & $M_3$ & $M_4$ \\ \hline
-0.1146 & -0.1336 & -0.1424 & -0.0528 & \textbf{0.0528} \\
0.2671 & 0.2785   & 0.5715   & 0.1699 & \textbf{0.1659} \\
1.0274 & 0.8569 & 2.0086 & 0.4576  & \textbf{0.4538}
\end{tabular} %3
& \begin{tabular}{ccccc} %4
$M_0$ & $M_1$ & $M_2$ & $M_3$ & $M_4$ \\ \hline
-0.1139 & -0.1662 & -0.2114 & -0.0948 & \textbf{0.0948} \\
0.3594 & 0.3987 & 0.7139 & 0.2384 & \textbf{0.1758} \\
1.3594 & 1.2600 & 2.8986 & 0.6208  & \textbf{0.4528}
\end{tabular} %4
\\ \hline
\end{tabular} %1
\end{center}
\end{table*}

\begin{table*}[htbp] \small
\caption{Performance metrics for model selection based on 160 testing images for each dataset. The models selected by each method are bolded. For the dataset MNIST-D$i$ (or FasionMNIST-D$i$), the associated ground truth model is $M_i$, $i=0,1,\ldots,4$.}
\label{table:stat_test}
\begin{center}
\begin{tabular}{c|c|c} %1
\hline
 & MNIST-D0 & MNIST-D1\\
\begin{tabular}{c} %2
 \\ \hline CRC \\ ER \\ CE
\end{tabular} %2
& \begin{tabular}{ccccc} %3
$M_0$ & $M_1$ & $M_2$ & $M_3$ & $M_4$ \\ \hline
\textbf{0.01} & -0.01 & -0.08  & -0.03     & -0.02 \\
0.0187 & \textbf{0.0125}   & 0.0400   & \textbf{0.0125} & \textbf{0.0125} \\
0.04 & 0.022 & 0.18 & \textbf{0.016}  & 0.03
\end{tabular} %3
& \begin{tabular}{ccccc} %4
$M_0$ & $M_1$ & $M_2$ & $M_3$ & $M_4$ \\ \hline
-0.1542 & \textbf{0.0886} & -0.1194  & -0.0886  & -0.0974 \\
0.0437 & \textbf{0}  & 0.0437   & 0.0187 & 0.0187 \\
0.1012 & \textbf{0.0035} & 0.1074 & 0.0493  & 0.0404
\end{tabular} %4
\\ \hline
 & MNIST-D2 & MNIST-D3\\
\begin{tabular}{c} %2
  \\ \hline CRC \\ ER \\ CE
\end{tabular} %2
& \begin{tabular}{ccccc} %3
$M_0$ & $M_1$ & $M_2$ & $M_3$ & $M_4$ \\ \hline
-0.2463 & -0.1046 & \textbf{0.1046}  & -0.1826  & -0.1452 \\
0.0938 & 0.0938 & \textbf{0.0063}  & 0.0750 & 0.0750 \\
0.5124 & 0.5020 & \textbf{0.0305} & 0.2375  & 0.2386
\end{tabular} %3
& \begin{tabular}{ccccc} %4
$M_0$ & $M_1$ & $M_2$ & $M_3$ & $M_4$ \\ \hline
-0.0733 & \textbf{0.0112} & -0.0704  & -0.0112  & -0.0167 \\
0.0375 & \textbf{0.0063} & 0.0250  & 0.0125 & 0.0187 \\
0.1239 & \textbf{0.0162} & 0.0531 & 0.0401  & 0.0276
\end{tabular} %4
\\ \hline
 & MNIST-D4 & FasionMNIST-D0\\
\begin{tabular}{c} %2
  \\ \hline CRC \\ ER \\ CE
\end{tabular} %2
& \begin{tabular}{ccccc} %3
$M_0$ & $M_1$ & $M_2$ & $M_3$ & $M_4$ \\ \hline
 -0.1727 & -0.0023 & -0.0651  & -4$\times 10^{-4}$  & \textbf{4$\times 10^{-4}$} \\
0.0750 & 0.0250 & 0.0625  & \textbf{0.0187} & 0.0250  \\
0.2656 & 0.0562 & 0.1678 & 0.0588  & \textbf{0.0484}
\end{tabular} %3
& \begin{tabular}{ccccc} %4
$M_0$ & $M_1$ & $M_2$ & $M_3$ & $M_4$ \\ \hline
-0.0625 & \textbf{0.0625} & -0.1348 & -0.1745 & -0.1533 \\
\textbf{0.1250} & 0.1313 & 0.4875 & 0.2250 & 0.2437 \\
\textbf{0.3760} & 0.4111 & 1.8613 & 0.6854  & 0.9154
\end{tabular} %4
\\ \hline
 & FasionMNIST-D1 & FasionMNIST-D2\\
\begin{tabular}{c} %2
  \\ \hline CRC \\ ER \\ CE
\end{tabular} %2
& \begin{tabular}{ccccc} %3
$M_0$ & $M_1$ & $M_2$ & $M_3$ & $M_4$ \\ \hline
-0.1126 & \textbf{0.0974} & -0.0974 & -0.1615 & -0.1696 \\
0.1250 & \textbf{0.1187} & 0.4000 & 0.2375 & 0.2625 \\
0.4188 & \textbf{0.3270} & 1.1290 & 0.6700  & 0.8646
\end{tabular} %3
& \begin{tabular}{ccccc} %4
$M_0$ & $M_1$ & $M_2$ & $M_3$ & $M_4$ \\ \hline
-0.2290 & -0.2385 & \textbf{0.2290} & -0.4017 & -0.2887 \\
0.6125 & 0.6250 & \textbf{0.1125} & 0.7188 & 0.7125 \\
2.0249 & 2.1765 & \textbf{0.3341} & 2.3835  & 3.1427
\end{tabular} %4
\\ \hline
 & FasionMNIST-D3 & FasionMNIST-D4\\
\begin{tabular}{c} %2
  \\ \hline CRC \\ ER \\ CE
\end{tabular} %2
& \begin{tabular}{ccccc} %3
$M_0$ & $M_1$ & $M_2$ & $M_3$ & $M_4$ \\ \hline
-0.1087 & -0.1353 & -0.1181 & -0.0629 & \textbf{0.0629} \\
0.2562 & 0.2500 & 0.5813 & \textbf{0.1562} & \textbf{0.1562} \\
0.9680 & 0.7373 & 1.8535 & 0.4849  & \textbf{0.4227}
\end{tabular} %3
& \begin{tabular}{ccccc} %4
$M_0$ & $M_1$ & $M_2$ & $M_3$ & $M_4$ \\ \hline
-0.0843 & -0.1795 & -0.1828 & -0.0597 & \textbf{0.0597} \\
0.3626 & 0.3688 & 0.7188 & 0.1938 & \textbf{0.1688}\\
1.2670 & 1.2215 & 2.7638 & 0.6143  & \textbf{0.4669}
\end{tabular} %4
\\ \hline
\end{tabular} %1
\end{center}
\end{table*}
We use a CNN architecture with 7 layers, namely, the input layer, the convolution layer, an average pooling layer, the flatten layer, 2 fully connected layers, and a softmax layer. In the convolution layer, 20 9-by-9 sized filters are used. The dimensions of the fully connected layers are 360 and 60, respectively. The type of activation function is ReLU. We apply a momentum based stochastic gradient descent algorithm for minimizing the cross-entropy loss, wherein we set the initial learning rate to be 0.1, then halve it after each training epoch; we set the momentum factor to 0.95, the minibatch size to 128, and the total epoch number to 3.

Based on the above setting, we evaluate the performance of the proposed CRC-based method by comparing it with conventional methods that do model selection according to the cross-entropy (CE) loss and the error rate (ER). We do not include NLL here, because it is in essence equivalent to CE. Note that both CE and ER based methods require access to labels of the testing images, while our CRC based method does not need any label to work.
\subsection{Experimental results}\label{sec:exp_res}
As described above, given a set of testing data points from the $i$th dataset, $M_i$ is our expected answer for the model selection problem, as $M_i$ is built by training a CNN model to fit the training data of the $i$th dataset, $i=0,\ldots,4$.
We do experiment to test whether the our CRC based method can identify the right model from the candidate models as well as its competitors.

We present the experimental results in Table I, in which ``CRC" denotes the CRC score. Note that, among these metrics, only the CRC score's computation does not need any data label.
We find that our CRC-based method selects the same model as the others for datasets MNIST-D0, MNIST-D1, MNIST-D2, MNIST-D3, FasionMNIST-D1, FasionMNIST-D2, FasionMNIST-D3 and FasionMNIST-D4. It performs better than the others for MNIST-D4, as it selects the right model, while the others all select a wrong one. It fails to identify the right model for FasionMNIST-D0, while the others perform well therein. To summarize, our label-free CRC based method performs comparably to the other label-based methods.

In the above experiments, we use all testing images with a sample size of $10^4$. To test the efficiency of the our CRC based method, we check whether reducing the sample size could degrade its performance. We set the sample size at 160 and present the corresponding results in Table II. As is shown, the CRC based method is the only one that succeeds in selecting the right model for MNIST-D0.
For MNIST-D1, MNIST-D2, MNIST-D3, FasionMNIST-D1, FasionMNIST-D2, FasionMNIST-D3, and FasionMNIST-D4, it performs as well as the others. For MNIST-D4, the CRC and CE based methods select the right model, while using ER leads to a wrong selection. Only for FasionMNIST-D0, the CRC based method performs worse than the others.
To summarize, for the above case with a small number of test images, our label-free CRC based method still performs comparably to conventional label based methods.
\section{Conclusions and Future Works}\label{sec:discussion}
Existent approaches to CNN model selection all depend on data labels. Here we investigate how to do LFMS for CNN based image recognition. As we know, confidence acts as an internal judgement for decision making, while it does not indicate the correctness of decisions. There is no clear relationship between confidence and the correctness of decisions. Here we show that, if we complement confidence with three other elements, namely relativity, consistency, and rationality, then the resulting CRC score can act as a reliable metric for measuring correctness. In particular, we show that, despite being extremely simple to implement, our CRC based LFMS method performs as well as conventional label data based methods.

Currently our method assumes that the ground truth model is included in the candidate model set. An interesting direction following this work is to investigate approaches to deal with cases wherein the ground truth model is not included in the candidate model set. A possible solution is to assign each candidate model a prior probability and then take the CRC score as a likelihood, in the similar spirit of \cite{pawitan2021confidence}, then employ Bayesian model averaging (BMA) \cite{domingos2000bayesian,maddox2018fast} to make predictions. Another interesting question that worths future investigation is how to do CNN model selection in dynamic settings. A possible solution is resorting to an algorithmic framework termed Bayesian dynamic ensembling of multiple models, a dynamic extension of BMA in principle \cite{liu2021robust2}.
\bibliographystyle{IEEEtran}
\bibliography{mybibfile}
\end{document}